\documentclass{article}

\usepackage[preprint]{neurips_2025}

\usepackage[utf8]{inputenc} 
\usepackage[T1]{fontenc}    
\usepackage{hyperref}       
\usepackage{url}            
\usepackage{booktabs}       
\usepackage{amsfonts}       
\usepackage{nicefrac}       
\usepackage{microtype}      
\usepackage{xcolor}         
\usepackage{wrapfig}
\usepackage{graphicx}
\usepackage{caption}

\usepackage{color}
\usepackage{graphicx}
\usepackage{multirow}
\usepackage{booktabs}
\usepackage{colortbl} 
\usepackage{amssymb}
\usepackage{caption}
\usepackage{color}
\usepackage{booktabs}    
\usepackage{graphicx}    
\usepackage{caption}     
\usepackage{array}       

\usepackage{amsmath}
\usepackage{amssymb}
\usepackage{algorithm} 
\usepackage[algo2e,ruled,linesnumbered]{algorithm2e}
\SetKwInOut{Parameter}{parameter}
\usepackage{enumitem}
\usepackage{marvosym}
\usepackage{ifsym}

\title{Controlled Data Rebalancing in Multi-Task Learning for Real-World Image Super-Resolution}

\author{%
  Shuchen Lin\\
  Xidian University\\
  \And
  Mingtao Feng\thanks{Corresponding author.}\\
  Xidian University\\  
  \And
  Weisheng Dong\\
  Xidian University\\
  \And
  Fangfang Wu\\
  Xidian University\\
  \And
  Jianqiao Luo\\
  Hunan University\\
  \And
  Yaonan Wang\\
  Hunan University\\
  \And
  Guangming Shi\\
  Xidian University\\
}

\begin{document}

\maketitle

\begin{abstract}

Real-world image super-resolution (Real-SR) is a challenging problem due to the complex degradation patterns in low-resolution images. Unlike approaches that assume a broadly encompassing degradation space, we focus specifically on achieving an optimal balance in how SR networks handle different degradation patterns within a fixed degradation space. We propose an improved paradigm that frames Real-SR as a data-heterogeneous multi-task learning problem. Our work addresses task imbalance in the paradigm through coordinated advancements in task definition, imbalance quantification, and adaptive data rebalancing. Specifically, we introduce a novel task definition framework that segments the degradation space by setting parameter-specific boundaries for degradation operators, effectively reducing the task quantity while maintaining task discrimination. We then develop a focal loss based multi-task weighting mechanism that precisely quantifies task imbalance dynamics during model training. Furthermore, to prevent sporadic outlier samples from dominating the gradient optimization of the shared multi-task SR model, we strategically convert the quantified task imbalance into controlled data rebalancing through deliberate regulation of task-specific training volumes. Extensive quantitative and qualitative experiments demonstrate that our method achieves consistent superiority across all degradation tasks. 
\end{abstract}

\vspace{-2mm}
\section{Introduction}
\label{sec:intro}
\vspace{-2mm}

Image super-resolution aims to recover high-resolution (HR) images from degraded low-resolution (LR) inputs. Unlike conventional non-blind~\cite{dong2014learning, zhang2018residual, zhang2018image} and blind~\cite{huang2020unfolding, wang2021unsupervised, gu2019blind} SR methods assuming simplified degradations, real-world SR (Real-SR) must contend with complex, unknown degradations, such as sensor noise\cite{zhang2021designing}, motion blur~\cite{bistron2024optimization}, compression artifacts~\cite{wang2024vcisr}, and optical aberrations~\cite{zhang2021designing}. These distortions corrupt high-frequency details and vary spatially, making traditional approaches inadequate for demanding applications~\cite{zhu2020csrgan,liebel2016single,wan2020deep}. Consequently, Real-SR frameworks must accommodate diverse, unpredictable degradation patterns while preserving perceptual fidelity.

To bridge the simulation-to-reality gap, prior studies have prioritized the design of sophisticated degradation models to synthesize training pairs that mimic real-world conditions. Pioneering works like RealESRGAN~\cite{wang2021real} employ randomized degradation pipelines combining degradation operators like blur, noise, resize and compression in varying sequences.
Such methods train a single SR network on synthetically degraded data, aiming for the broadest coverage of potential degradation cases. 

On the other hand, our work shifts focus to examining the consistency of SR network optimization across different degradation patterns within a fixed degradation space. TGSR~\cite{zhang2023real} has posited that the rebalancing of diverse degradation patterns can be fundamentally framed as a multi-task learning paradigm with data heterogeneity, where each task corresponds to a set of HR-LR image pairs generated under a specific degradation pattern. For convenience, we will hereinafter also refer to such a degradation pattern as a 'task'. A straightforward and basic strategy proposed by TGSR is to discretize the degradation space by sampling a large set of degradation parameter combinations, and treating each as a separate task to form a task set $\mathcal{T}=\left\{\tau_i\right\}_{i=1}^N$. To mitigate task imbalance from the multi-task learning perspective, the task grouping algorithm proposed by TGSR fine-tunes a shared model using all sampled degradation tasks, and then clusters these tasks based on absolute individual performance gains, thereby aiming to break through performance bottlenecks on unsatisfactory tasks. 

\begin{figure*}[t]  
    \centering
    \includegraphics[width=1\textwidth]{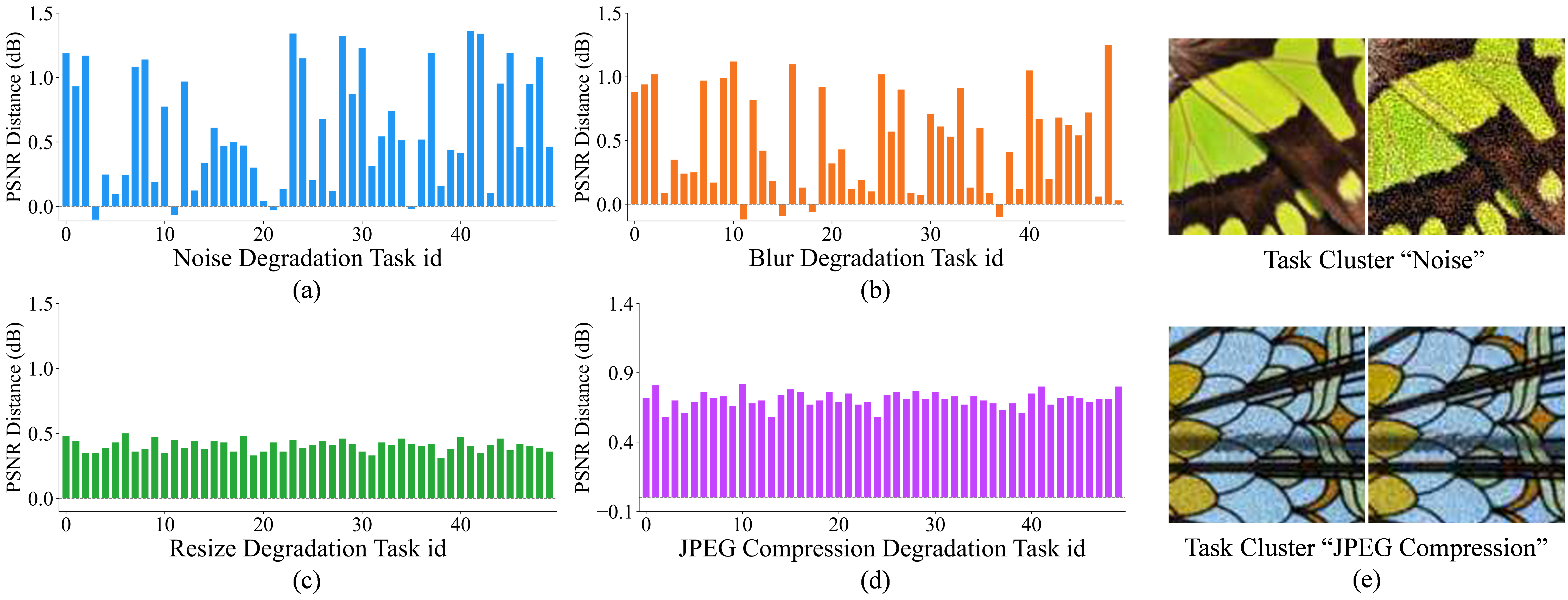}
    \vspace{-5mm}
    \caption{Task-defining capacity of each degradation operator.} 
    \label{val_exp}
\vspace{-10mm}
\end{figure*}

In this paper, we delve further into this paradigm. We observe that not all degradation operators exhibit the same strong discriminative power when defining tasks, some operators contribute redundantly to task differentiation. To empirically analyze the discriminative capacity of individual degradation operators, we conduct a validation experiment based on the degradation pipeline in~\cite{wang2021real}. The results in Fig.~\ref{val_exp} reveal that operators such as noise injection and blur artifacts exert significantly stronger influence on task formulation and imbalance, demonstrating a higher discriminative power compared to others like down-sampling or JPEG compression. We thus propose simplifying the paradigm definition, focusing primarily on the more impactful degradation operators as the defining factors for distinct tasks, thereby deprioritizing those less influential in task differentiation. Concurrently, we partition the degradation space by the parameter ranges of these influential operators, redefining tasks as sets of sample pairs from each resulting subspace. 

Building upon an explicit definition of degradation tasks, we systematically investigate task imbalance within the multi-task SR paradigm and identify two critical issues. (1) Tasks possess various inherent difficulties. Methods relying on absolute performance gains (like \cite{zhang2023real}) wrongly assume that equal gains reflect equivalent task competition, without accounting for inherent task difficulties. (2) Learning progress is dynamic. Both task imbalance and relative learning progress evolve significantly throughout the training process. To address these challenges, we introduce a dynamic multi‑task loss‑weighting algorithm that quantifies task imbalance by referencing single‑task model performance to implicitly capture difficulty. Furthermore, our method adapts to the evolving training dynamics by periodically re‑evaluating and updating loss weights throughout training.

Previous studies~\cite{gregoire2024sample} have demonstrated that naive multi-task loss weighting can destabilize optimization by overweighting outliers. To address this stability challenge, we establish that under our proposed task definition framework, the weighting term of multi-task losses is equivalent to controlling the sample quantity for each task. Consequently, we strategically regulate task-specific sample generation at different intervals, address task imbalance through controlled data rebalancing. Notably, the conventional practice of constructing LR training images from HR counterparts through degradation models enables straightforward manipulation of task-specific data quantities. Our findings reveals new potential for multi-task learning paradigms in real-world SR scenarios. 

\textbf{Our main contributions are:} \textbf{(1)} We reframe the paradigm of multi-task learning for real-world image super-resolution by proposing a novel task definition framework. This definition removes redundant degradation parameters while preserving degradation model's capacity. \textbf{(2)} We develop a difficulty-aware loss weighting mechanism to dynamically quantify and address task imbalance during training. \textbf{(3)} We theoretically prove that quantified task imbalance equivalently corresponds to adjustable data quantity imbalance, providing an optimization stabilization strategy. Specifically, this strategy can be readily integrated with other Real-SR methods.

\vspace{-2mm}
\section{Related Works}
\vspace{-2mm}
\noindent\textbf{Real-world super-resolution.}
Compared to non-blind SR methods, Real-SR requires handling complex degradations. Early works~\cite{kim2016accurate, ledig2017photo, zhang2018learning} model degradation spaces using Gaussian kernels and noise. BSRGAN~\cite{zhang2021designing} and RealESRGAN~\cite{wang2021real} advance explicit degradation modeling via shuffled and high-order strategies. Recently, diffusion-based models~\cite{wu2024one, wu2024seesr, fan2024adadiffsr} address intricate degradations via iterative refinement but incur high inference costs. Somewhat similarly, DASR~\cite{wang2021unsupervised} proposes three degradation patterns and introduces an additional network for degradation recognition. In contrast, our method prioritizes balanced training progression across distinct degradation patterns within a fixed degradation space, rather than expanding it to accommodate broader degradation variations.

\noindent\textbf{Multi-task learning.}
Multi-task learning methodologies can generally be classified into three main types: (1) Task rebalancing~\cite{guo2018dynamic, kendall2018multi} mitigates task imbalance via loss weighting or gradient manipulation. (2) Task grouping~\cite{zamir2018taskonomy, fifty2021efficiently} identifies synergistic tasks for joint learning, leveraging concepts like task taxonomy~\cite{zamir2018taskonomy} or affinity scores~\cite{fifty2021efficiently}. (3) Architecture design methods include hard parameter sharing (shared encoder, distinct decoders)~\cite{kokkinos2017ubernet} and soft parameter sharing (separate networks with cross-talk)~\cite{misra2016cross}. Our approach adopts the most relevant task rebalancing methods to quantify and address the issue of degradation task imbalance within SR models.

\vspace{-2mm}
\section{Method}
\vspace{-2mm}
\textbf{Problem Formulation.} Real-SR addresses the ill-posed inverse problem of reconstructing a high-resolution (HR) image $x \in \mathbb{R}^{H \times W \times c}$ from its degraded low-resolution (LR) observation $y \in \mathbb{R}^{h \times w \times c}$, where the degradation process $\mathcal{D}(\cdot)$ explicitly encapsulates the complexities inherent in physical imaging systems. Formally, the forward degradation model can be expressed as:
\begin{equation}
\ y=\mathcal{D}(x ; \Theta)=\left(f_{\theta_n} \circ \cdots \circ f_{\theta_2} \circ f_{\theta_1}\right)(x)
\end{equation}
Here, $\mathcal{D}(x;\Theta)$ is a parameterized degradation model with a set of empirically predefined parameters $\Theta=\{\theta_n, \cdots, \theta _2, \theta_1\}$, each function $f_{\theta_i}$ represents an individual degradation operator, such as blurring, noise addition, down-sampling, or JPEG compression, applied in a sequential manner. Existing Real-SR methods (e.g. RealESRGAN~\cite{wang2021real}) customize $\mathcal{D}$ through the binding of stochastic sampling parameters and optimize the sequential ordering to mimic real imaging pipelines.

We recall the conception proposed by TGSR, mapping the Real-SR onto a multi-task learning problem. Since the degradation model $\mathcal{D}$ can be regarded as a vast degradation space, an SR task $\tau$ is defined as training pairs $(X, Y=d(X))$, where the degradation $d(\cdot)$ is sampled from the infinite degradation space $\mathcal{D}$ and applied to an HR image set $X$ to produce an LR image set $Y$. Due to the infinite size of $\mathcal{D}$, TGSR chose to sampled a large number of fixed degradation parameters to form the task space, and tried to handle the task competition problem from a multi-task learning perspective.

\begin{figure*}[t]  
    \centering
    \includegraphics[width=1\textwidth]{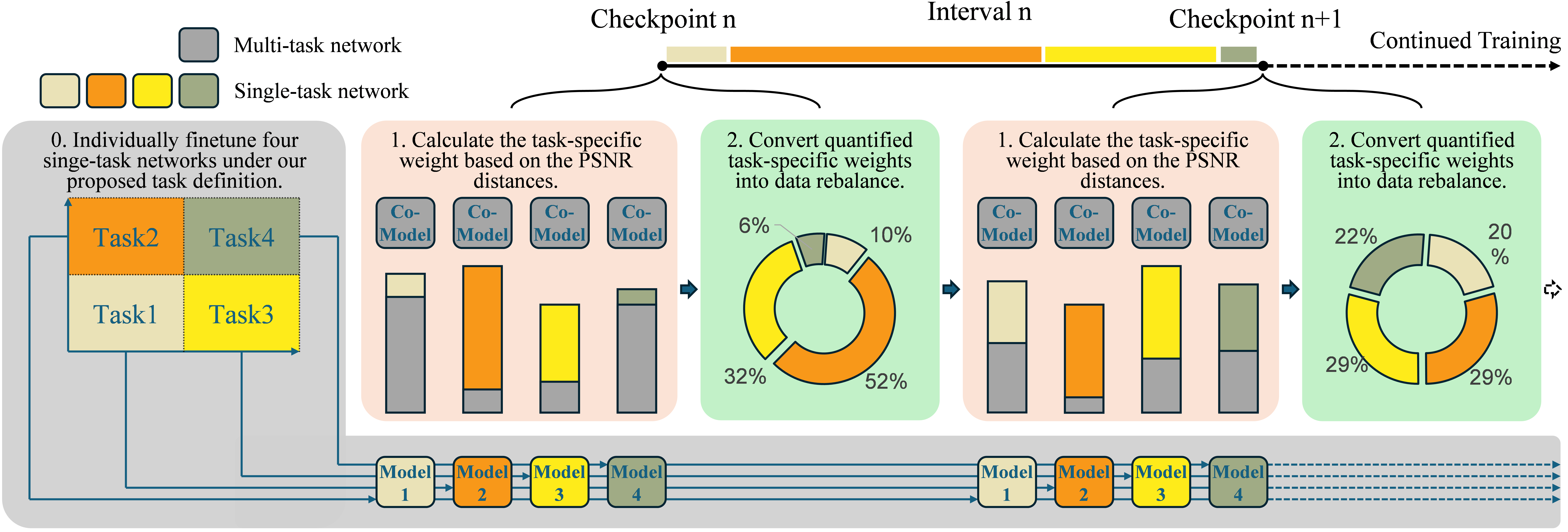}
    \vspace{-3mm}
    \caption{Overview of our proposed pipeline. At the beginning of each training interval, PSNR distances for each task are measured relative to the respective single-task networks to compute task weights, which are then converted into the task data volume for the duration of that interval.}
    \label{overview}
    \vspace{-6mm}
\end{figure*}

The pipeline of our framework is shown in Fig.~\ref{overview}. The detailed introduction of each part is below.

\subsection{Task Redefination via Degradation Subspace Partitioning}
To quantitatively validate the task discrimination ability of different degradation operators, we conduct a validation experiment. For each operator (e.g., Gaussian blur, noise, JPEG compression and down-sampling), we create 50 degradation configurations by varying only the severity of the target operator while keeping others fixed. We then fine-tune a pre-trained Real-SR model on each configuration separately, yielding 50 operator-specific single-task models. The PSNR differences between the pre-trained model and each fine-tuned model are used to assess the task-specific learning progress. A larger variance in PSNR improvements indicates stronger task imbalance. As illustrated in Fig.~\ref{val_exp}(a) and Fig.~\ref{val_exp}(b), noise and blur operators exhibit substantial variance, indicating severe task imbalance during joint training of shared models. In stark contrast, Fig.~\ref{val_exp}(c) and Fig.~\ref{val_exp}(d) reveal a markedly stable PSNR distances variation curve while varying down-sampling (and JPEG compression) degradation parameters, implying weaker task imbalance under this set of configurations. 

The results in Fig.~\ref{val_exp} show that different degradation operators $f_{\theta_i}$ demonstrate heterogeneous contribution weights in task formulation, as their representational capacities vary significantly across the degradation space $\mathcal{D}$. In other words, degradation task clusters formed by varying intensity levels of noise contamination operators (and blur artifacts operators) demonstrate significantly stronger task imbalance and negative transfer problems compared to those formed by other degradation operators (e.g., down-sampling or JPEG compression artifacts). 

Building upon these insights,  we strategically shift our task definition focus to noise contamination operators and blur artifacts operators while excluding redundant hyperparameters (e.g. fixed down-sampling ratios or predetermined compression levels), thereby refining the task definition into a more practical and performance-impactful paradigm. Meanwhile, unlike~\cite{zhang2023real} samples a number of fixed degradation task parameters to represent the whole degradation space - this random and discrete task sampling approach inevitably compromises the expressiveness of the original degradation space $\mathcal{D}$ - we establish the parameter values of noise and blur operators as degradation task boundaries to partition the degradation space $\mathcal{D}$, redefine tasks as training data pairs generated from distinct continuous degradation subspaces.

\begin{wrapfigure}{l}{0.5\textwidth} 
    \centering
    \includegraphics[width=0.5\textwidth]{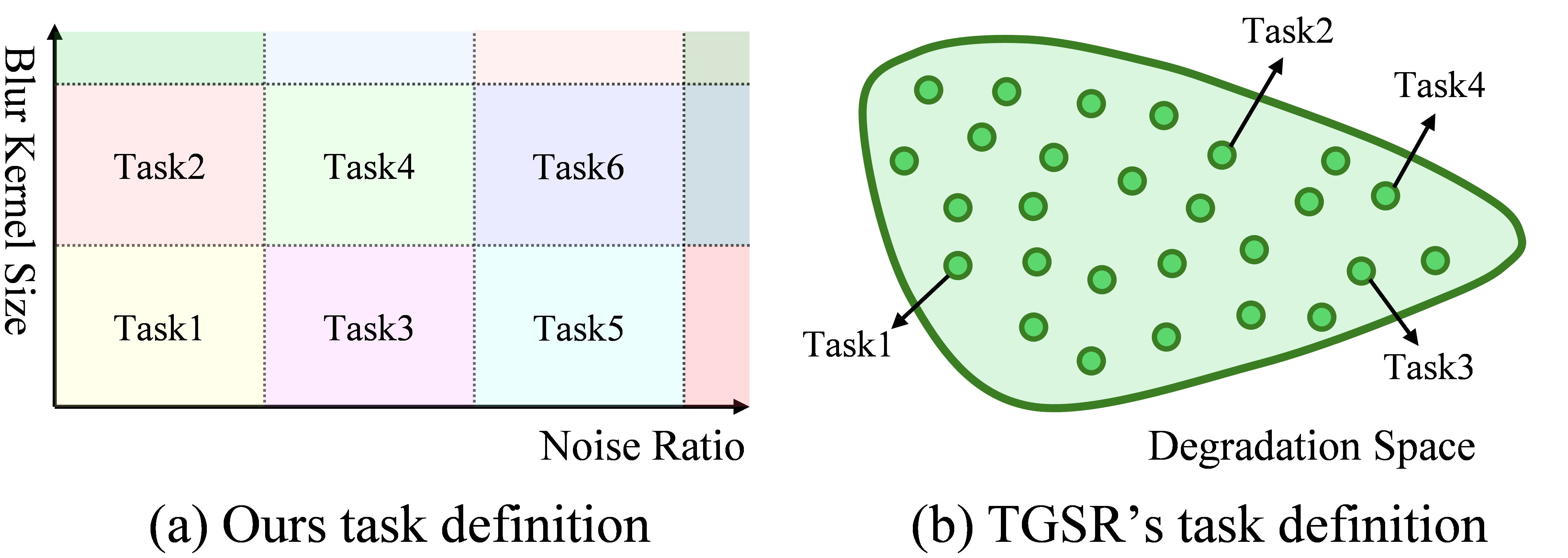}
    \vspace{-5mm}
    \caption{Divergent task definition between our method and TGSR.}
    \label{task_define}
    \vspace{-4mm}
\end{wrapfigure}

Specifically, we partition the degradation space $\mathcal{D}$ into $n$ distinct subspaces based on the parametric values of noise and blur operators, $\mathcal{D}_{\text {sub}}=\left\{\mathcal{D}_1, \mathcal{D}_2, \ldots, \mathcal{D}_n\right\}$. Given the whole set of HR images $\mathcal{X}$ ($|\mathcal{X}| \gg n$), we construct $n$ SR tasks $\mathcal{T}=\left\{\tau_i=\left(X_i, Y_i\right)\right\}_{\mathrm{i}=1}^{\mathrm{n}}$, where each $X_i \in \mathcal{X}$ corresponds to an LR counterpart $Y_i$ generated by applying degradation parameters from the subspace $\mathcal{D}_i\in\mathcal{D}_{\text {sub}}$. Therefore, without compromising the expressive power of degradation models, Real-SR is reformulated as a multi-task learning framework, where a shared model collaboratively addresses degradation tasks derived from distinct degradation subspaces. Fig.~\ref{task_define} illustrates the differences in task definitions between our method and TGSR.

\noindent\textbf{Analysis for Task Imbalance.}
From the multi-task learning perspective, task imbalance occurs when tasks exhibit uneven learning progress or disparate resource allocation during joint optimization, leading to compromised convergence or sub-optimal performance for individual tasks~\cite{yu2020gradient}. For instance, in tasks combining semantic segmentation and depth estimation, gradients from one task may dominate parameter updates, suppressing the learning of other tasks.

In the context of conventional multi-task learning, task imbalance stems primarily from task diversity, reflecting the absence of adequately shared latent representations across different tasks. On the other hand, within our task definition framework, the involved tasks are semantically homogeneous as they all  belong to the super-resolution domain, yet exhibit heterogeneity in their degradation prior distributions. As we illustrated in Fig.~\ref{val_exp}(a) and (b), the degradation tasks partitioned by varying severity levels may compete for the shared capacity of jointly trained models. This competition could induce an optimization bias where the model preferentially learns to restore textures with higher-frequency distortions (e.g. severely degraded regions requiring intensive reconstruction), while insufficiently addressing the restoration needs of mildly degraded images.

Most real-world SR algorithms utilize randomized degradation parameters to generate LR training data. This unbiased generation approach is equivalent to giving all degradation tasks equal optimization priority. However, when the entire task cluster exhibits severe task imbalance, owing to lack of external regulation, such random task selection mechanism tends to impede the optimization efficiency of shared models and ultimately restricts the performance of under-optimized tasks.

\subsection{Adaptive Loss Weighting for Multi-Task Super-Resolution}
Loss function re-weighting is an effective external regulatory mechanism to address task imbalance in multi-task field. In this section, we present our methodology for quantifying task imbalance to determine adaptive loss weights. As analyzed in Sec.~\ref{sec:intro}, our loss weighting strategy must account for two critical aspects: (1) inherent task difficulty discrepancies arising from varying degradation severity levels, (2) dynamic evolution of task imbalance throughout the training process. Drawing inspiration from ~\cite{yun2023achievement}, we formulate our adaptive SR task weighting scheme based on the focal loss.

Focal loss was proposed to address class imbalance in dense detection scenarios, and it modifies the standard cross-entropy loss by introducing a focusing weighting term $\left(1-p_t\right)$ to prioritize hard-classified examples. The loss function is formulated as:
\begin{equation}
F L\left(p_t\right)=\left(1-p_t\right)^\gamma C E=-\left(1-p_t\right)^\gamma \log \left(p_t\right)
\end{equation}
where $p_t$ denotes the model's estimated probability for the true class. The focusing weighting term $\left(1-p_t\right)$ dynamically scales loss contributions by suppressing well-classified samples while preserving higher weights for ambiguous cases, forcing optimization to prioritize challenging examples.

Building upon the focusing weighting term, we define the loss weight for task $\tau$ as follows:
\begin{equation}
w_\tau=\operatorname{DP}\left(N_{\text {single }}^\tau, N_{\text {multi }}^\tau\right)
\label{psnr_distance}
\end{equation}
where $\operatorname{DP}\left(N_{\text {single }}^\tau, N_{\text {multi }}^\tau\right)$denotes the PSNR distance on validation set generated by degradation task $\tau$ between the dedicated single-task network trained on task $\tau$ and multi-task network built on joint training. Compared with solely relying on the absolute PSNR values of the shared model, this relative measurement mitigates the impact of inherent task difficulty variations. Tasks exhibiting smaller PSNR distances are identified as dominant competitors in current shared training process, and we accordingly assign them lower weights to balance gradient updates across tasks.

However, since the shared model serves as a mediator for cross-task knowledge transfer during joint training, the validation PSNR of the shared model may surpass that of the single-task network, resulting in negative PSNR distances. To address this phenomenon, we reformulate the task weights through applying an exponential transformation.

Meanwhile, to further accommodate the evolving task imbalance dynamics during joint optimization, we introduce temporal partitioning of the training schedule into $K$ intervals, $T=\left\{T_1, T_2, \ldots,T_k, \ldots,T_K\right\}$, each spanning a fixed number of iterations. At the onset of every interval $T_k$, we dynamically calibrate all task weights by recomputing the exponential weighting formula given by Equ.~\ref{psnr_distance} using the latest validation metrics. After the final weight normalization, the proposed multi-task SR loss can be formulated as:
\begin{equation}
\begin{aligned}
&\text { For each } T_k \in T, \quad \mathcal{L}_{\mathrm{SR}}\left(T_k\right)=\sum_{\tau \in \mathcal{T}} w_\tau^{k} \mathcal{L}_\tau \\
&w_\tau^{k}=\exp \left(\operatorname{DP}\left(N_{\text {single }}^\tau, N_{\text {multi }}^\tau\right)\right)
\label{final loss}
\end{aligned}
\end{equation}

\subsection{Controlled Data Rebalancing for Addressing Task Imbalance}
\label{data imbalance}
Although our loss function weighting strategy successfully modulates task imbalance in multi-task super-resolution learning, recent studies~\cite{gregoire2024sample} reveal that naive application of loss weighting may cause unintended amplification of outlier impacts. To prevent this phenomenon, we propose to redirect the quantified task imbalance through the perspective of sample prior distribution. By establishing multi-task weights as the bridge, our framework strategically converts the data rebalancing into a mechanism for compensating task imbalance.

We begin from Equ.~\ref{final loss} and reformulate the multi-task loss function into a sample re-weighting formulation. Given a set of degradation tasks $\mathcal{T}=\left\{\tau_{i}=\left(X_{i}, Y_{i}\right)\right\}_{i=1}^{n}$, each task corresponds to a degradation subspace $\mathcal{D}_{\text {sub}}=\left\{\mathcal{D}_1, \mathcal{D}_2, \ldots, \mathcal{D}_n\right\}$. Within a specific interval $T_k$ (determined by a fixed number of iterations), the total number of training samples is $N$. We first assume that all tasks share an identical sample size, i.e., $N_1=N_2 \ldots=N_n=N / n$. The loss function for task $\tau_i$ is defined as follows:
\begin{equation}
\mathcal{L}_{\tau_i}=\frac{1}{N_i} \sum_{(x_i, y_i) \in \tau_i} \ell(x_i, y_i)
\end{equation}
where $\ell(x_i, y_i)$ denotes the single sample loss. Substituting this into the multi-task loss function Equ.~\ref{final loss} yields:
\begin{equation}
\mathcal{L}=\sum_{i=1}^n w_{\tau_i}^{k} \cdot \frac{1}{N_i} \sum_{(x_i, y_i) \in \tau_i} \ell(x_i, y_i)
\end{equation}
By interchanging the order of summation, we derive a unified loss summation over all samples:
\begin{equation}
\mathcal{L}=\sum_{i=1}^n \sum_{(x_i, y_i) \in \tau_i} \frac{ w_{\tau_i}^{k}}{N_i} \ell(x_i, y_i)
\end{equation}
We define the sample weight $w(x_i, y_i)$ as:
\begin{equation}
w(x_i, y_i)=\frac{w_{\tau_i}^{k}}{N_i}, \quad \text { for }(x_i, y_i) \in \tau_i
\end{equation}
Thus, the loss function can be reformulated as:
\begin{equation}
\mathcal{L}=\sum_{(x, y) \in \mathcal{T}} w(x, y) \ell(x, y)
\end{equation}
Through this formulation, we convert the multi-task weighted loss into sample re-weighting, where samples from task $\tau_i$ are assigned weight of ${w_{\tau_i}^{k}}/{N_i}$.

Next, we enforce equal sample weights across all tasks, while adjusting the sample size of each task to preserve the task imbalance quantification results from Section 3.2. Specifically, we set the sample weight $w(x_i, y_i)$ to a uniform value:
\begin{equation}
\forall \tau_i \in \mathcal{T}, \quad w\left(x_i, y_i\right)=\frac{w_{\tau_i}^k}{N_i}=\alpha
\end{equation}
We substitute $\sum_{i=1}^n N_i=N$ and $\sum_{i=1}^n w_{\tau_i}^k=1$ into the above equation and solve for $\alpha$:
\begin{equation}
\alpha=\frac{\sum_{i=1}^n w_{\tau_i}^k}{N}=\frac{1}{N}
\end{equation}
Consequently, the adjusted sample size for each task in interval $T_k$ is:
\begin{equation}
N_i=\frac{w_{\tau_i}^k}{\alpha}={N \cdot w_{\tau_i}^k}
\end{equation}
Through the above transformations, we equivalently convert the quantified task imbalance weights into controllable data rebalancing, thereby preventing outlier values with large weights from excessively dominating optimization, stabilizing model training.
\section{Experiment}
\noindent\textbf{Training Datasets.}
We employ DIV2K~\cite{agustsson2017ntire}, Flickr2K~\cite{agustsson2017ntire} and OutdoorSceneTraining~\cite{wang2018recovering} datasets for training. We use the same degradation pipeline as RealESRGAN~\cite{wang2021real} to synthesize LR-HR pairs.

\noindent\textbf{Test Datasets.}
For evaluation, we construct the DIV2K4Level dataset from the DIV2K validation set, comprising four distinct validation subsets aligned with the four degradation subspaces defined in our task formulation. Each subset consists of 100 image pairs. Beyond this, we evaluate our model on a synthetic test set, DIV2K-Val, as well as two real-world datasets, RealSR~\cite{cai2019toward} and DRealSR~\cite{wei2020component}. The synthetic dataset comprises 100 image pairs. The LQ images are synthesized by applying random degradations using Real-ESRGAN~\cite{wang2021real} to DIV2K\_valid~\cite{agustsson2017ntire}.

\noindent\textbf{Evaluation Metrics.}
For evaluating our method, we apply both fidelity and perceptual quality metrics. Fidelity metrics include PSNR and SSIM~\cite{wang2004image} (calculated on the Y channel in YCbCr space). Perceptual quality metrics include LPIPS~\cite{zhang2018unreasonable}.

\noindent\textbf{Implementation details.}
We adopt the high-order degradation model proposed by RealESRGAN~\cite{wang2021real} as the Real-SR degradation model in our experiments. We partitioned the degradation space into 4 subspaces, termed mild, blur, noise, and severe, based on the value ranges of two parameters: blur kernel size and noise injection rate. Detailed degradation settings and other implementation details can be found in the supplementary material. 

\vspace{-2mm}
\subsection{Comparison with State-of-the-Art}
\vspace{-2mm}
We compare our method with the state-of-the-art methods, contrastive learning method include DASR~\cite{liang2022efficient}, Gan-based methods include RealESRGAN~\cite{wang2021real}, SwinIR~\cite{liang2021swinir}, and MM-RealSR~\cite{mou2022metric}. Diffusion-based methods include ResShift~\cite{yue2023resshift}, SinSR~\cite{wang2024sinsr}, TSD-SR~\cite{dong2024tsd}, and AdcSR~\cite{chen2024adversarial}. Officially released pre-trained models are used for the compared methods.

\noindent\textbf{Quantitative Comparisons on DIV2K4Level. }DASR defines three degradation types for image degradation estimation. It achieves exceptionally high PSNR metrics, which is often attributed to specific hyperparameter configurations in its loss function weighting. However, its performance on LPIPS is notably poor (the worst and exhibiting a significant disparity).  In contrast, MM-RealSR and RealESRGAN utilize higher-order degradation models. These approaches typically sacrifice some PSNR performance to achieve enhanced perceptual quality. SwinIR, which leverages the Swin Transformer architecture for image super-resolution, demonstrates strong performance across selected evaluation metrics. Diffusion-based methods generally excel on non-reference metrics, indicating their capacity to generate perceptually realistic images. However, the outputs from these models can sometimes exhibit considerable differences from the ground truth images, consequently leading to lower performance on reference-based metrics. The result is shown in Tab.~\ref{DIV2K4Level}. Our proposed method consistently demonstrates either the best or second-best performance across all degradation settings, underscoring the effectiveness of our balanced approach to these degradation tasks.

\begin{table*}[t]
\centering
\renewcommand{\arraystretch}{1.2}
\resizebox{\linewidth}{!}{
\begin{tabular}{l|ccc|ccc|ccc|ccc}
\toprule[1pt]
\multicolumn{1}{c|}{\multirow{2}{*}{}} & \multicolumn{3}{c|}{DIV2K\_mild}                 & \multicolumn{3}{c|}{DIV2K\_blur}                 & \multicolumn{3}{c|}{DIV2K\_noise}                         & \multicolumn{3}{c}{DIV2K\_severe}                                \\
\multicolumn{1}{c|}{\multirow{2}{*}{}} & PSNR                 & LPIPS & SSIM                & PSNR                 & LPIPS & SSIM                 & PSNR                 & LPIPS & SSIM                 & \multicolumn{1}{c}{PSNR} & LPIPS & \multicolumn{1}{c}{SSIM} \\ \hline
DASR(CVPR'21)~\cite{liang2022efficient}                 & \underline{23.72}                     &0.5581       & \underline{0.6508}                     & 23.40                     & 0.6466      &     0.5677                 &    \underline{23.68}                  & 0.6403      &    0.5782                  &   23.46                       & 0.6344      & 0.5863                          \\
RealESRGAN(ICCV'21)~\cite{wang2021real}                &23.62  &0.4273       &0.6382  &23.44  &0.4343       &0.6314  &23.58  &\underline{0.4330}      & 0.6364 & 23.39                         &0.4403       & 0.6292                          \\
RealSwinIR(ICCV'21)~\cite{liang2021swinir}           & 23.52                     & 0.4372      & 0.6374                     &23.46                      & 0.4416      & 0.6337                     & 23.45                     &0.4424       &    0.6337                  &   23.39                       &  0.4471     &0.6299                           \\ 
MM-RealSR(ECCV'22)~\cite{mou2022metric}             &  23.44                    & 0.4293      & 0.6468                     & 23.45                     & 0.4352      &  \underline{0.6430}                    & 23.44                     &0.4343       &  \textbf{0.6448}                    & 23.45                         & 0.4412      &\underline{0.6340}                           \\\hline
ResShift(NIPS'23)~\cite{yue2023resshift}   &  23.71           &   0.4646                   & 0.6174      &  23.41                    & 0.4711                     &  0.6110     &    23.57                  &  0.4728                    &0.6090       & \underline{23.48}                     &    0.4796                      & 0.6031                               \\
SinSR(CVPR'24)~\cite{wang2024sinsr}              &23.39                      &0.4835      &  0.5654                   & 23.29                   & 0.4901      &   0.5585                   &23.24                      &0.4938       &  0.5524                    &  23.13                        & 0.5004      &  0.5456                         \\
TSD-SR(CVPR'25)~\cite{dong2024tsd}               &22.16                      &  \underline{0.4256}    &0.5753                      &  22.04                    & \textbf{0.4308}      &  0.5694                    &  22.16                    & 0.4380      & 0.5727                     &  22.03                        & \textbf{0.4333}      &   0.5666                        \\
AdcSR(CVPR'25)~\cite{chen2024adversarial}              & 23.27                     & 0.4272     &  0.6050         &23.18                      &\underline{0.4332}       &0.6002                     & 23.22                     & 0.4336      &   0.6000                   &  23.13                        & 0.4398      &   0.5949                        \\ \hline
TGSR(NIPS'23)~\cite{zhang2023real}              &23.64            &0.4283       &0.6425                      &\underline{23.49}                      & 0.4354      &  0.6359	                     & 23.54                     &0.4339       &0.6395                      & 23.39                         & 0.4415      & 0.6330                          \\
Ours                 &\textbf{23.79}  &\textbf{0.4190}       &\textbf{0.6511}  &\textbf{23.66}  &\textbf{0.4308}       &\textbf{0.6483}  &\textbf{23.73}  & \textbf{0.4272}      &\underline{0.6424}  & \textbf{23.54}                       & \underline{0.4383}     &\textbf{0.6396}            \\
\bottomrule[1pt]
\end{tabular}
}
\vspace{-2mm}
\caption{Quantitative results of different methods on DIV2K4Level.}
\label{DIV2K4Level}
\vspace{-5mm}
\end{table*}

\begin{table}[t]
\centering
\renewcommand{\arraystretch}{1.2}
\resizebox{\linewidth}{!}{
\begin{tabular}{l|clc|clc|clc|clc}
\toprule[1pt]
\multicolumn{1}{c|}{\multirow{2}{*}{}} & \multicolumn{3}{c|}{RealSRset-Nikon}                 & \multicolumn{3}{c|}{RealSRset-Canon}                 & \multicolumn{3}{c|}{DRealSR}                         & \multicolumn{3}{c}{DIV2K-Val}                                \\
\multicolumn{1}{c|}{\multirow{2}{*}{}} & PSNR                 & LPIPS & SSIM                & PSNR                 & LPIPS & SSIM                 & PSNR                 & LPIPS & SSIM                 & \multicolumn{1}{c}{PSNR} & LPIPS & \multicolumn{1}{c}{SSIM} \\ \hline
DASR(CVPR'21)~\cite{liang2022efficient}                 &\textbf{27.15}                      &0.4391       & 0.7598                     &\textbf{27.67}                     &0.4236       &  0.7893                    & \textbf{30.41}                     & 0.4387      & \underline{0.8261}                   & \textbf{24.43}                        & 0.5726      &      \underline{0.6438}                     \\
RealESRGAN(ICCV'21)~\cite{wang2021real}           &25.62                     &0.3820       &0.7607                      &26.06                      &0.3629       & 0.7864                     &28.64                      & 0.3766     &0.8052                      &23.50                          &0.4341       &0.6338                          \\
RealSwinIR(ICCV'21)~\cite{liang2021swinir}           &25.72                 &\underline{0.3768}      & \underline{0.7663}                   &26.33                      &\textbf{0.3573 }      & 0.7945                     & 28.24                     & 0.3819      & 0.7983                     &  23.45                        & 0.4424      &  0.6335                         \\ 
MM-RealSR(ECCV'22)~\cite{mou2022metric}            &23.54                      &0.3822       & 0.7425                     &  24.06                    &0.3606       & 0.7708                     &     26.97                 &\underline{0.3723}     & 0.7972                     &23.43                          & 0.4353      & 0.6434                          \\ \hline
ResShift(NIPS'23)~\cite{yue2023resshift}              & 25.11                     &0.4736       & 0.6999                     & 25.87                     & 0.4626      & 0.7501                     & 27.12                     & 0.4689      &  0.7407                    &  23.69                       & 0.4724      &  0.6101                         \\
SinSR(CVPR'24)~\cite{wang2024sinsr}              &25.68                      & 0.4702    &  0.6843                    &  26.35                    &0.4596       & 0.7279                     & 28.45                     & 0.4496      & 0.7516                     &   23.25                       & 0.4921      & 0.5552                          \\
TSD-SR(CVPR'25)~\cite{dong2024tsd}               & 23.71                     & 0.4023      &  0.6849                    & 23.89                     &  0.3956     &  0.7073                    &26.23                     &0.4019       &0.7160                      &   22.10                       &\underline{0.4338}      & 0.5709                          \\
AdcSR(CVPR'25)~\cite{chen2024adversarial}              &25.49                      & 0.3925      &0.7187                      & 25.71                    &0.3796       & 0.7457                     &  28.21                    &0.3822       & 0.7723                     & 23.20                         &  0.4343     & 0.5996                          \\ \hline
TGSR(NIPS'23)~\cite{zhang2023real}                 &25.88  &0.3862       &0.7688  &26.35  &0.3660       &\underline{0.7954 } &28.91  &0.3839       &0.8142  & 23.51                         &0.4348       &0.6376                           \\
Ours                 &\underline{26.16}  &\textbf{0.3763}      &\textbf{0.7728}  &\underline{26.54}  &\underline{0.3580}       &\textbf{0.7960}  &\underline{29.27}  &\textbf{0.3697}      &\textbf{0.8393}  &\underline{23.72}                         &\textbf{0.4254}      &\textbf{0.6441}   \\
\bottomrule[1pt]
\end{tabular}
}
\vspace{1mm}
\caption{Quantitative Comparisons on Synthetic and Real-World Test Sets. }
\label{Real-World test results}
\vspace{-8mm}
\end{table}

\begin{figure*}[t]  
    \centering
    \includegraphics[width=1\textwidth]{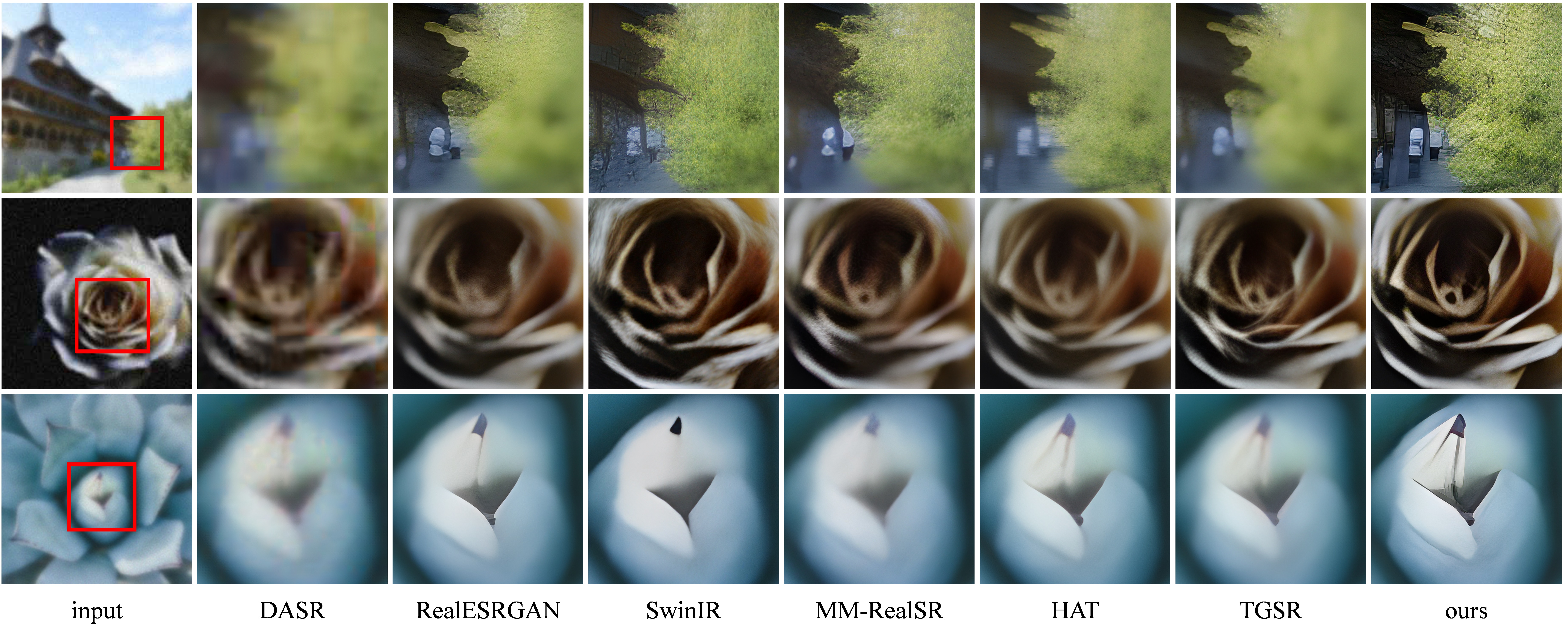}
    \vspace{-6mm}
    \caption{Qualitative results of different methods. Zoom in for details.}
    \label{Qualitative Comparisons}
    \vspace{-4mm}
\end{figure*}

\noindent\textbf{Quantitative Comparisons on Synthetic and Real-World Test Sets. }
Our method demonstrates consistent superiority not only on specialized task-specific benchmarks but also across the synthetic DIV2K-Val dataset and real-world test sets (RealSRset \& DRealSR). The result is shown in Tab.~\ref{Real-World test results}. These results demonstrate that our method does not merely achieve a performance balance among the delineated degradation tasks, but rather fulfills our foundational goal: achieving an optimal balance in how SR networks handle different degradation patterns within a fixed degradation space. 

\noindent\textbf{Qualitative Comparisons.}
The qualitative comparisons in Fig.~\ref{Qualitative Comparisons} demonstrate the superiority of our framework over mainstream Real-SR methods. Under heavy and mixed degradations (first two rows), competing approaches either fail to remove blur and noise or introduce unnatural textures, whereas our model effectively restores structures with minimal artifacts. In high‑contrast regions (last row), our results exhibit crisper edges and more faithful texture reconstruction compared to the overly smooth or artifact‑ridden outputs of prior methods.

\subsection{Analysis and Ablation Study}
\begin{figure}[t]  
    \centering
    \includegraphics[width=1\textwidth]{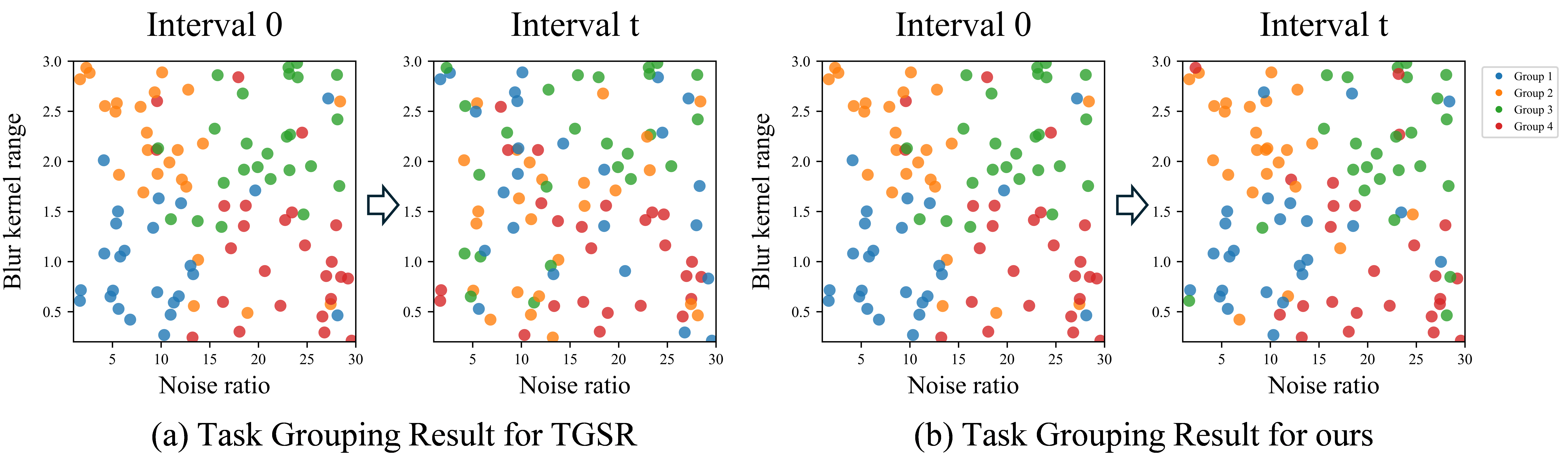}
    \vspace{-5mm}
    \caption{Illustration of the grouping results of 100 selected tasks for TGSR and our method, where tasks grouped into same groups are represented with identical colors. }
    \label{task_grouping}
    \vspace{-4mm}
\end{figure}

\noindent\textbf{Task Imbalance Dynamics Analysis during Training.}
Fig.~\ref{task_grouping} visualizes the evolution of task imbalance during model training. We selected two checkpoints during the training process, and employed grouping indicator from TGSR to evenly group 100 randomly sampled discrete tasks into 4 groups. The resulting changes in grouping outcomes revealed how task imbalance evolved. Fig.~\ref{task_grouping}(a) follows the TGSR framework, where task grouping is executed once at initialization and remains fixed throughout training. The grouping outcomes in the two intervals differ markedly, indicating that task imbalance and training progression vary substantially over the course of training. In contrast, Fig.~\ref{task_grouping}(b), which adopts our proposed method, reveals that the clustering results display stronger consistency. These observations demonstrate that while task imbalance undergoes significant shifts during TGSR training (due to its static grouping strategy), our method dynamically captures and adapts to these imbalances via adaptive weight updates. In specific experiments, we set parameter t to 40, corresponding to half of the complete training procedure.

\begin{figure}[t]
    \centering
    \begin{minipage}{0.53\textwidth}
        \centering
        \renewcommand{\arraystretch}{1.2}
        \scriptsize{}
        \begin{tabular}{l@{\hspace{23pt}}llll}
        \toprule[1pt]
        \multicolumn{1}{c}{} & \multicolumn{2}{c}{RealSRset-Nikon} & \multicolumn{2}{c}{RealSRset-Canon} \\
        \multicolumn{1}{c}{} & PSNR           & LPIPS           & PSNR           & LPIPS           \\ 
        \hline
        RealSRGAN~\cite{ledig2017photo}            & 24.71               & 0.4159                 &  25.42               & 0.3902                  \\
        RealSRGAN+ours      & \textbf{25.27}               & \textbf{0.3997 }              &  \textbf{25.86 }            & \textbf{0.3752}              \\ \hline
        RealESRGAN~\cite{wang2021real}           &25.62           &0.3820           &26.06           & 0.3629 \\
        RealESRGAN+ours     &\textbf{26.16}                &\textbf{0.3763}                 &\textbf{26.54 }               &\textbf{0.3580 }                \\ \hline
        RealSwinIR~\cite{liang2021swinir}           & 25.72               & 0.3768                &  26.33               & 0.3573   \\
        RealSwinIR+ours     & \textbf{26.68  }             & \textbf{0.3652}             &  \textbf{27.01        }      & \textbf{0.3415}                \\ \hline
        RealHAT~\cite{chen2023hat}              & 27.17               & \textbf{0.3612 }               & 27.72               &  \textbf{0.3360 }              \\
        RealHAT+ours      &\textbf{27.65}                &   0.3691              &  \textbf{28.01 }             & 0.3417                \\ 
        \bottomrule[1pt]
        \end{tabular}
        \captionof{table}{The quantitative results of applying proposed framework to mainstream Real-SR methods.}
        \label{integrate result}
    \end{minipage}%
    \hfill
    \begin{minipage}{0.43\textwidth}
        \centering
        \includegraphics[width=1.0\textwidth]{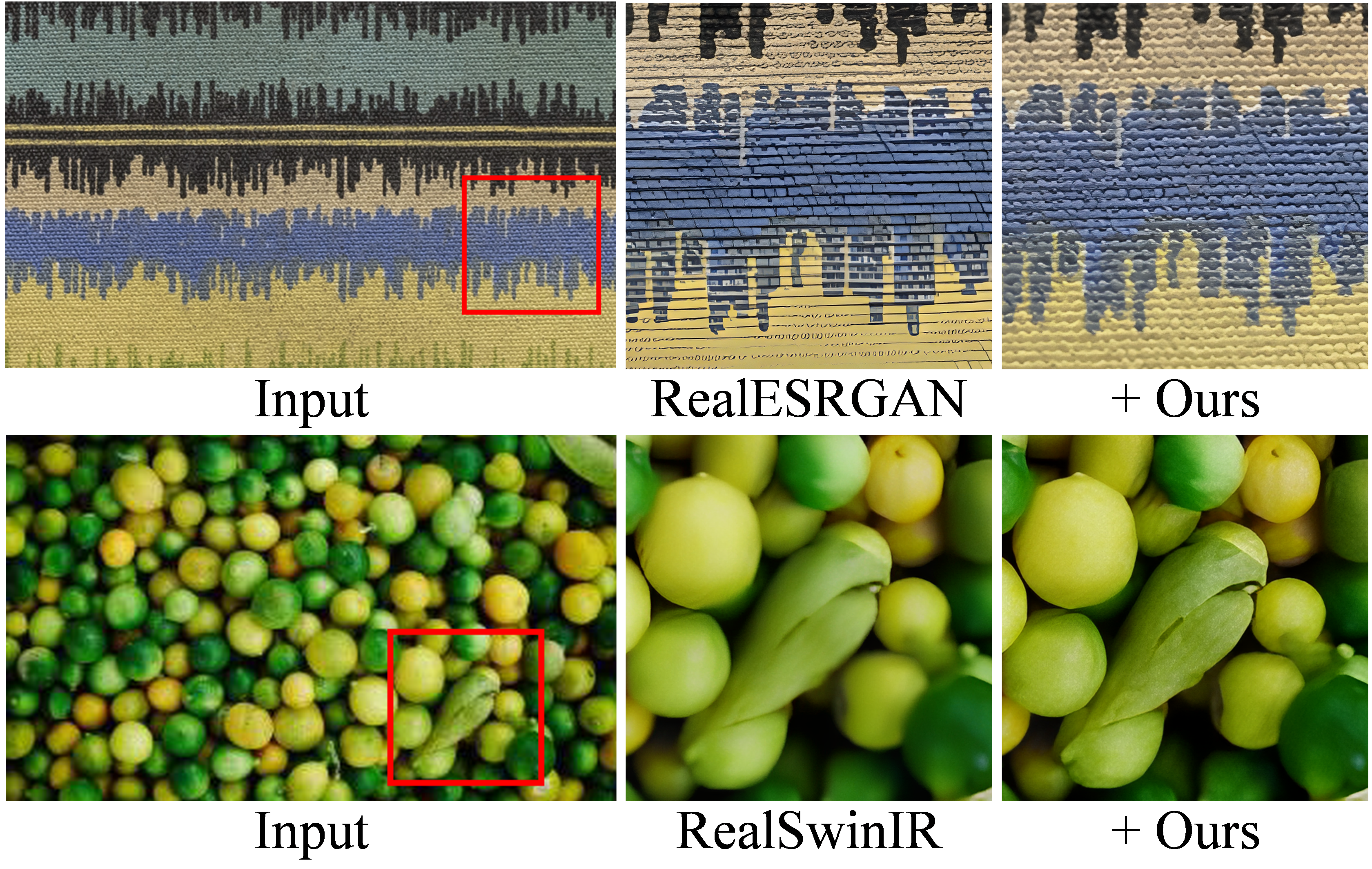}
        \vspace{-15pt}
        \caption{The qualitative results of algorithm integrability.}
        \label{integrate vision result}
    \end{minipage}
    \vspace{-3mm}
\end{figure}

\noindent\textbf{Algorithm Integrability Verification. }
To validate the algorithmic integrability of our method, we conduct extensive experiments by integrating our framework into mainstream real-world image super-resolution models. As our approach builds a multi-task learning framework based on training data heterogeneity modeling, it imposes no assumptions on the specific architecture of the base SR model. We select 4 mainstream models—RealSRGAN~\cite{ledig2017photo}, RealESRGAN~\cite{wang2021real}, SwinIR~\cite{liang2021swinir}, and HAT~\cite{chen2023hat}—as integration baselines. Without modifying their original architectures or training configurations, we apply our task definition framework and dynamic loss weighting mechanism to their training pipelines. As shown in Tab.~\ref{integrate result}, our method achieves significant performance gains across real-world datasets, Fig.~\ref{integrate vision result} shows the visual improvements of our method compared to the baselines, for regions with composite degradations and complex textures, baseline methods tend to uniformly smear details, whereas our method demonstrates superior detail recovery.

\begin{figure}[t]
\centering
\begin{minipage}{0.53\textwidth}
    \centering
    \begin{minipage}{\textwidth}
        \centering
        \renewcommand{\arraystretch}{1.2}
        \scriptsize
        \begin{tabular}{cc|ccccc}
            \toprule[1pt]
            \multirow{2}{*}{Diff} & \multirow{2}{*}{Dyna} & \multirow{2}{*}{Methods} & \multicolumn{2}{c}{RealSRset-Nikon} & \multicolumn{2}{c}{RealSRset-Canon} \\
            & & & PSNR & LPIPS & PSNR & LPIPS \\ \cline{3-7}
            \checkmark & & RLW & 26.02 & 0.3833& 26.14 &0.3699 \\
            & \checkmark & DWA & 25.82 & 0.3759 & 26.38 & 0.3594 \\
            \checkmark & \checkmark & GLS & 25.68 & 0.3801 & 26.10 & 0.3606 \\
            \checkmark & \checkmark & Ours & \textbf{26.16} & \textbf{0.3763} & \textbf{26.54} & \textbf{0.3580} \\
            \bottomrule[1pt]
        \end{tabular}
        \captionof{table}{Ablation study for loss weighting algorithms.}
        \label{tab:loss_weighting}
    \end{minipage}

    \vspace{3mm}

    \begin{minipage}{\textwidth}
        \centering
        \renewcommand{\arraystretch}{1.2}
        \scriptsize
        \begin{tabular}{c@{\hspace{13.3pt}}ccccc}
            \toprule[1pt]
            \multirow{2}{*}{} & \multirow{2}{*}{\#tasks} & \multicolumn{2}{c}{RealSRset-Nikon} & \multicolumn{2}{c}{RealSRset-Canon} \\ 
            & & PSNR & LPIPS & PSNR & LPIPS \\ \hline
            w/ quantity set & 2 &25.89 &0.3759 &26.29 &0.3580 \\
            w/ quantity set & 3 &26.02 &0.3783 &26.40 &0.3615 \\
            w/ quantity set & 5 &26.14 &\textbf{0.3749} &26.48 &\textbf{0.3552} \\ \hline
            w/o quantity set & 4 &25.83 & 0.3788&26.31 &0.3597  \\
            w/ quantity set & 4 & \textbf{26.16} & 0.3763& \textbf{26.54} & 0.3559 \\
            \bottomrule[1pt]
        \end{tabular}
        \captionof{table}{Ablation study for the quantity of tasks and data rebalancing strategy.}
        \label{tab:task_quantity}
    \end{minipage}
\end{minipage}%
\hfill
\begin{minipage}{0.43\textwidth}
    \centering
    \makebox[\linewidth][r]{%
        \includegraphics[width=1.02\linewidth]{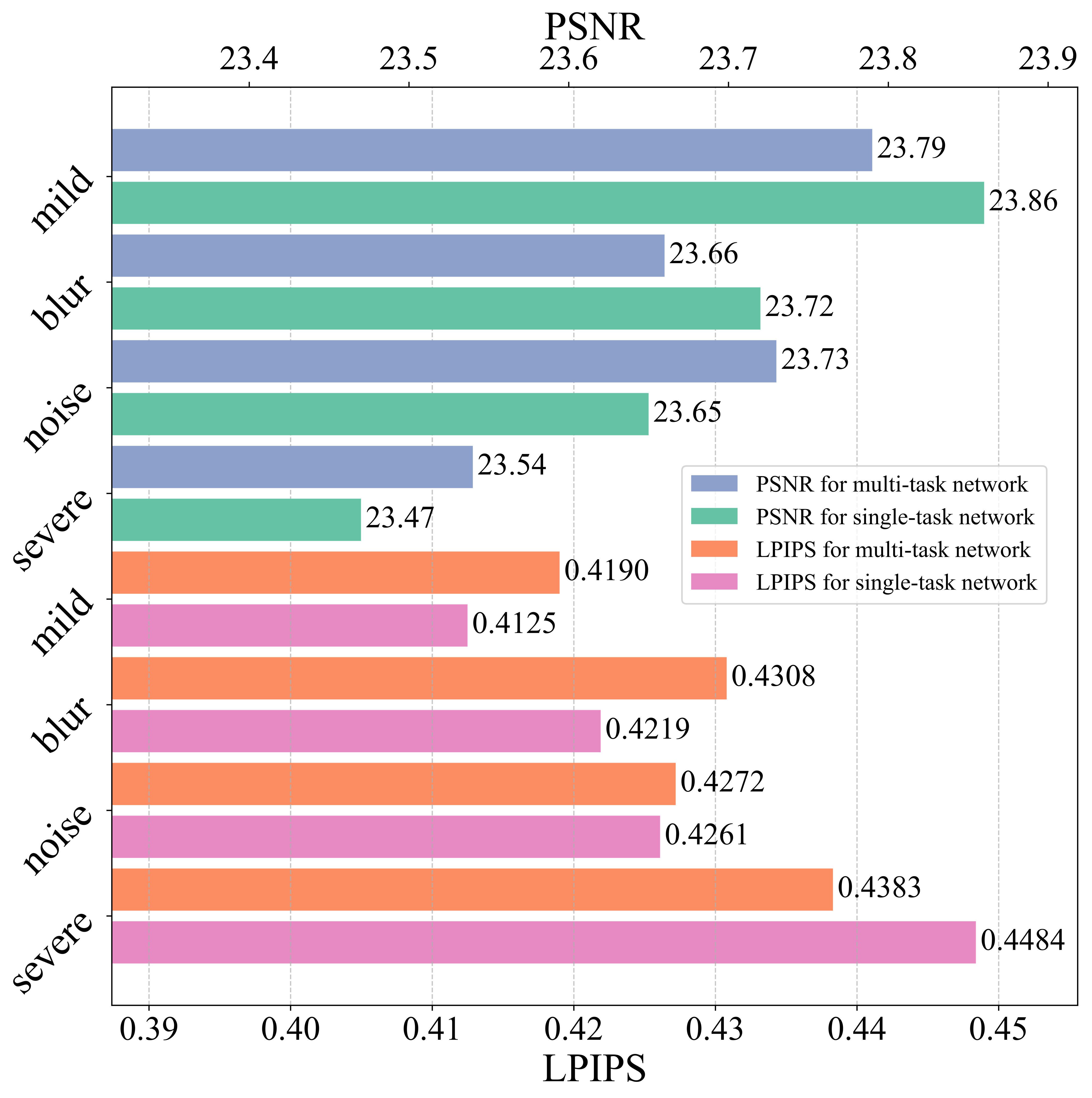}
    }
    \vspace{-19pt}
    \caption{Comparison of Single-task Networks and Proposed Multi-task Network.}
    \label{fig:upper_bound}
\end{minipage}
\vspace{-5mm}
\end{figure}

\noindent\textbf{Study of the performance upper bound. }
We train 4 dedicated single-task networks for each degradation subspace, and compare their performance against our multi-task network on respective DIV2K4Level validation sets. As shown in Fig.~\ref{fig:upper_bound}, our multi-task network achieved performance metrics comparable to those of the single-task networks across all degradation tasks. Notably, for the 'noise' and 'severe' degradation tasks, our multi-task network achieved PSNR improvements of 0.08 dB and 0.07 dB, respectively, surpassing the performance of the single-task models. These results highlight the capability of our framework to leverage cross-task information interaction, demonstrating its potential in addressing highly degraded real-world image super-resolution scenarios. 

\noindent\textbf{Ablation Study for Loss Function Weighting Algorithms. }
To investigate the effectiveness of our proposed loss weighting method in addressing task imbalance, we conduct an ablation study on the loss weighting component. We compare our approach with several classical multi-task loss weighting algorithms, including RLW~\cite{lin2021reasonable}, DWA~\cite{liu2019end} and GLS\cite{chennupati2019multinet++}. As shown in Tab.~\ref{tab:loss_weighting}, our task weighting algorithm achieves superior performance in real-world image super-resolution applications. This improvement can be attributed by the fact that the loss function weighting of traditional multi-task learning is mainly oriented to heterogeneous task semantics (e.g., cross-domain tasks like segmentation and depth estimation), while our proposed method aims to better handle task imbalance under different data distribution definitions.

\noindent\textbf{Other Ablation Study Results. }
We conducted supplementary ablation studies to validate the effectiveness of our task-imbalance-to-data-imbalance transformation approach described in Sec.~\ref{data imbalance}, and investigating model performance under varying granularities of task grouping configurations. As demonstrated in Tab.~\ref{tab:task_quantity}, our algorithm consistently achieves stable performance gains across different task quantity settings. Notably, while finer task partitioning yields finite incremental gains, it incurs linearly increasing training costs proportional to the number of task groups.

\vspace{-2mm}
\section{Conclusion}
\vspace{-2mm}
This work enhances multi-task real-world image super-resolution by refining task-space modeling and task imbalance management. We propose a degradation operator-aware task definition framework that segments the degradation space with parameter-specific boundaries, balancing task discrimination and efficiency. A focal loss-based weighting mechanism dynamically quantifies task imbalance, while a task-to-data imbalance conversion strategy stabilizes optimization by regulating task-specific training volumes. Experiments demonstrate consistent superiority across various degradation scenarios, with seamless integration into existing architectures. A potential limitation of our method is that the training cost increases proportionally with the number of defined tasks, this inherently restricts the extent to which tasks can be subdivided.

{
\small
\bibliography{main}
\bibliographystyle{plain}

}





\end{document}